\documentclass[english]{cccconf}
\usepackage[comma,numbers,square,sort&compress]{natbib}
\usepackage{epstopdf}
\usepackage{amsmath}
\usepackage{url}
\usepackage{color}

\begin{document}

\title{LLM-based Robot Task Planning with Exceptional Handling for General Purpose Service Robots}

\author{Ruoyu Wang$^{\ast}$ \aref{seu},
        Zhipeng Yang$^{\ast}$ \aref{seu},
        Zinan Zhao \aref{seu},
        Xinyan Tong \aref{seu},
        Zhi Hong \aref{seu},
        Kun Qian $^{\dagger}$ \aref{seu}}


\affiliation[seu]{School of Automation, Southeast University, Nanjing, 210096 \email{kqian@seu.edu.cn}}

\maketitle

\begin{abstract}
The development of a general purpose service robot for daily life necessitates the robot's ability to deploy a myriad of fundamental behaviors judiciously. Recent advancements in training Large Language Models (LLMs) can be used to generate action sequences directly, given an instruction in natural language with no additional domain information. However, while the outputs of LLMs are semantically correct, the generated task plans may not accurately map to acceptable actions and might encompass various linguistic ambiguities. LLM hallucinations pose another challenge for robot task planning, which results in content that is inconsistent with real-world facts or user inputs. In this paper, we propose a task planning method based on a constrained LLM prompt scheme, which can generate an executable action sequence from a command. An exceptional handling module is further proposed to deal with LLM hallucinations problem. This module can ensure the LLM-generated results are admissible in the current environment. We evaluate our method on the commands generated by the RoboCup@Home Command Generator, observing that the robot demonstrates exceptional performance in both comprehending instructions and executing tasks.

\end{abstract}

\keywords{Task Planning, General Purpose Service Robots, Large Language Model}

\footnotetext{$^\ast$ Authors with equal contribution.}
\footnotetext{$^\dagger$ Corresponding authors.}
\footnotetext{This work is supported by National Natural Science Foundation of China under Grant 61573101, Jiangsu Province Natural Science Foundation under Grant BK20201264 and Zhejiang Laboratory under Grant 2022NB0AB02.}

\section{Introduction}
The development of a general purpose service robot for daily life has long been an aspiration. A primary challenge in this task lies in enabling robots to comprehend the command in natural language and subsequently execute corresponding actions in a coherent sequence. This necessitates the robot's ability to deploy a series of fundamental behaviors judiciously, sorting and executing these behaviors under the overall task requirements. For instance, when presented with the command, \textit{Could you find a female person in the living room and answer her question?} the robot should infer a sequence of actions: first, navigate to the living room, then identify a female individual, and finally, respond to her question. 

Traditionally, these challenges have been approached from the perspectives of lexical analysis\cite{misra2016tell,misra2015environment}. However, high-level task planning often demands the definition of extensive domain knowledge related to the robot's operating environment and understanding of semantic knowledge. Autoregressive large language models (LLMs) trained on extensive corpora precisely address these requirements. Recent advancements in training Large Language Models (LLMs) have empowered systems to generate intricate texts based on prompts, answer questions, and engage in conversations spanning a wide range of topics. This capability has been applied to generate acceptable action plans within the context of robotic task planning\cite{li2022pre,ahn2022can}, either by scoring the next steps or directly generating new steps. 

However, while the outputs of LLMs are semantically correct, the generated task plans may not accurately map to acceptable actions and might encompass various linguistic ambiguities~\cite{singh2023progprompt}. Prior works also introduce LLM hallucinations~\cite{ji2023survey,huang2023survey}, resulting in content that is inconsistent with real-world facts or user inputs, poses another challenge for robot task planning. Faithfulness hallucinations refers to the divergence of generated content from user instructions, and factuality hallucinations refers to factual inconsistency or fabrication.

\begin{figure}[t]
	\centering
	\includegraphics[width=9cm]{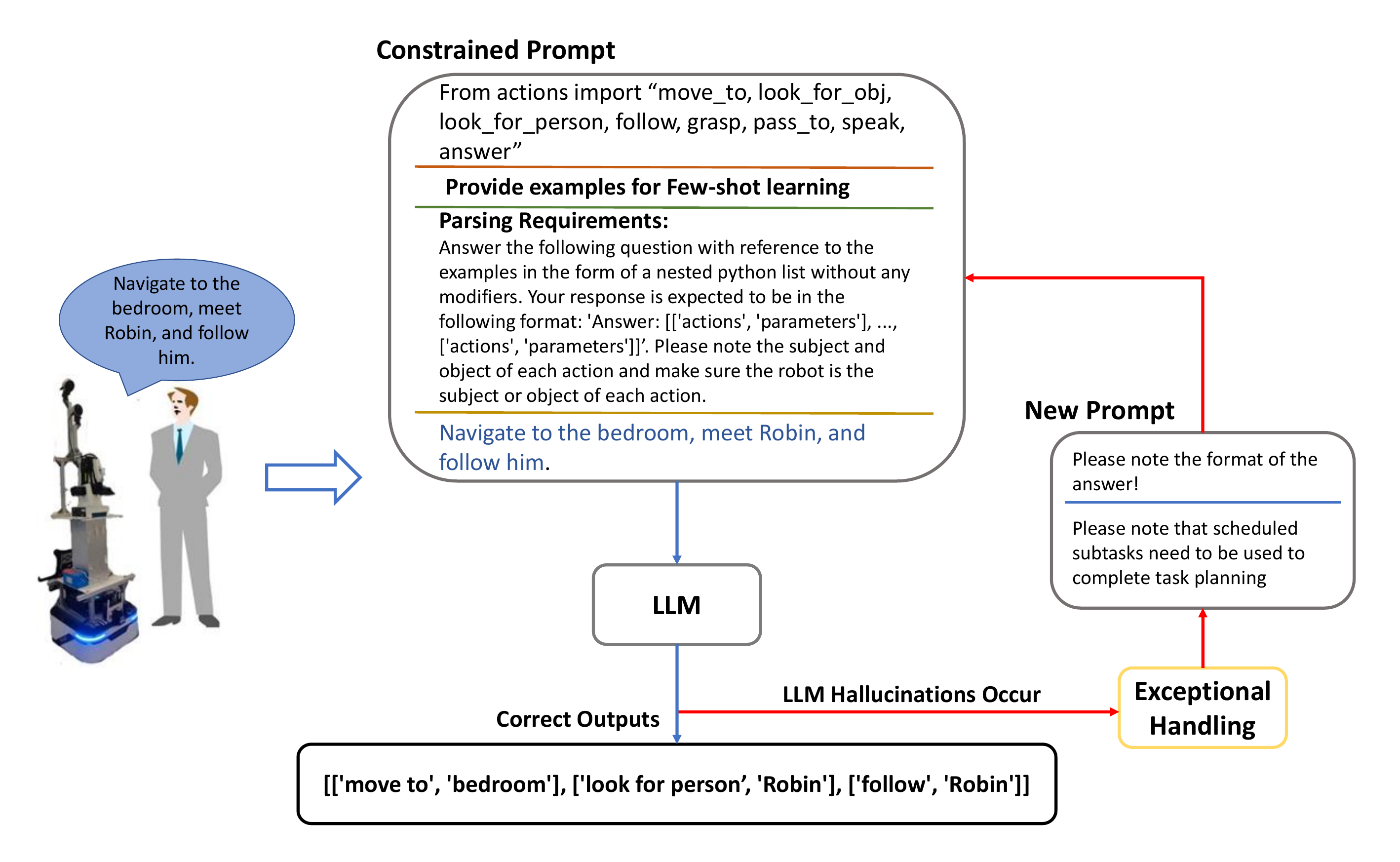}
	\caption{The work process of our prompt scheme.}
\end{figure}

\begin{figure*}[htbp]
	\centering
	\includegraphics[width=16cm]{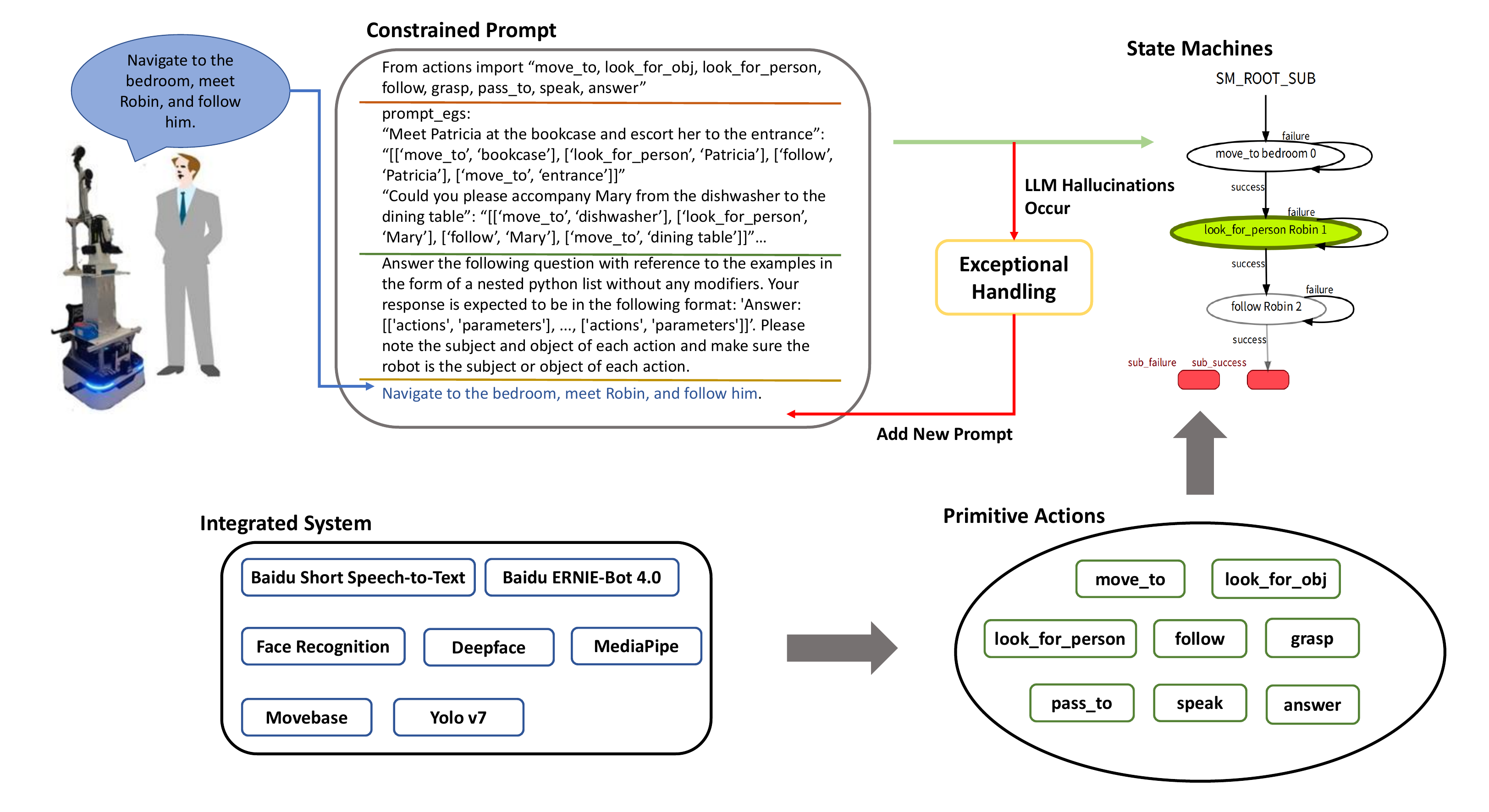}
	\caption{An overview of our method.}
	\label{chutian}
\end{figure*}

In light of this background, we propose a task planning method based on a constrained LLM prompt scheme, which can generate an executable action sequence from a command in natural language. An exceptional handling module is further proposed to deal with LLM hallucinations problem. This module can ensure the LLM-generated results are admissible in the current environment. Our strategy involves importing primitive actions as Pythonic functions and prompting the LLM to generate task planning results in the form of a nested Python list. These encourage the LLM to constrain the task planning result to executable primitive actions that are predefined. As for the exceptional handling module, we add new prompt instructions when we detect that the answer deviates from the specified format or the parsed function list contains actions beyond the provided primitive actions. Finally, we evaluate our method on the commands generated by the RoboCup@Home Command Generator~\cite{generator}, observing that the robot demonstrates exceptional performance in both comprehending instructions and executing tasks.

\section{Related Work}
Various prior works have looked into using language as a space for planning. In the beginning, a series of them\cite{misra2015environment,misra2016tell} applied lexical analysis to parse language instructions and resolve linguistic ambiguities, requiring many hand-designed rules to adapt to complex tasks and environments. Recently, since the remarkable ability large language models has shown in multitask generalization, many task planning approaches leverage pre-trained autoregressive LLMs to decompose abstract and high-level instructions into a sequence of low-level steps. 

Brown et al.\cite{brown2020language} showed that zero-shot or few-shot prompting could be highly effective for transfer learning. Furthermore, chain-of-thought prompting\cite{wei2022chain} proved that even simple prompting modifications can profoundly boost LLMs' performance. Li et al.\cite{li2022pre} incorporated environment context and predicted actions in VirtualHome\cite{puig2018virtualhome} by fine-tuning GPT-2. In contrast, Huang et al.\cite{huang2022language} investigated existing knowledge in LLMs (GPT-3\cite{brown2020language} and Codex\cite{chen2021evaluating}) to generate action plans for embodied agents, without any additional training.
 SayCan\cite{ahn2022can} applied a learned value function to evaluate each candidate action generated by LLM, which serves as a proxy for affordance. However, It may scale inadequately to dynamic environments with combinatorial action spaces. Inner Monologue\cite{huang2022inner} introduced grounded feedback from the environment to the LLM when generating each step in the plan and highly improved high-level instruction completion. PROGPROMPT\cite{singh2023progprompt} proposed a programmatic LLM prompt structure containing program-like specifications of the available actions and objects in an environment, which enables LLM to generate an entire, executable plan program directly. Obinata et al.\cite{obinata2023foundation} first utilized LLM(GPT-3\cite{brown2020language}) to perform General Purpose Service Robot (GPSR) tasks in robocup@home, getting high scores in the competition. However, it may not be robust in handling LLM hallucinationss~\cite{ji2023survey,huang2023survey} error. Based on this, We have incorporated exception handling and made compensations for weaknesses in our prompt design, resulting in a higher task completion rate.  
\section{Method}
\subsection{Method Overview}
Given a human command in daily life, our goal is to parse the command into an actions sequence and execute them sequentially. We formulate this task as follows:
\begin{equation}
    \mathbf{t_{n+1}}=f(\mathcal{K}, \mathcal{L}, \mathbf{t_{n}}, \mathbf{s})
\end{equation}
where $\mathbf{t_n} \in \mathcal{T}$ represents the robot action in the current environment. $\mathcal{K}$ is a set of all the objects available in the environment, and $\mathcal{L}$ is a set of executable actions that are predefined by the robot. $\mathbf{s}$ is the given command in natural language such as \textit{Give me a cola} and we need to construct a mapping function $f(\cdot)$ that generate available robot actions.

We propose a general purpose service robot system that integrates a highly executable task planning method with fundamental action models(see Fig. 2). Initially, we identified eight executable actions encompassing all conceivable scenarios arising from commands. Then we input the current command into LLM, which parses the command and generates an action sequence comprising the aforementioned eight primitive actions. Finally, the robot generates a state machine based on the action sequence and calls the encapsulated basic model to execute the state machine. To be more specific, we use Baidu Short Speech-to-Text for command input module, Baidu ERNIE-Bot 4.0~\cite{wenxin} for LLM, deepface for the face recognition module, MediaPipe~\cite{lugaresi2019mediapipe} for the gesture recognition and skeleton tracking and YOLO V7~\cite{wang2023yolov7} for the object detection model.

\begin{figure}[t]
  \centering
  \includegraphics[width=8cm]{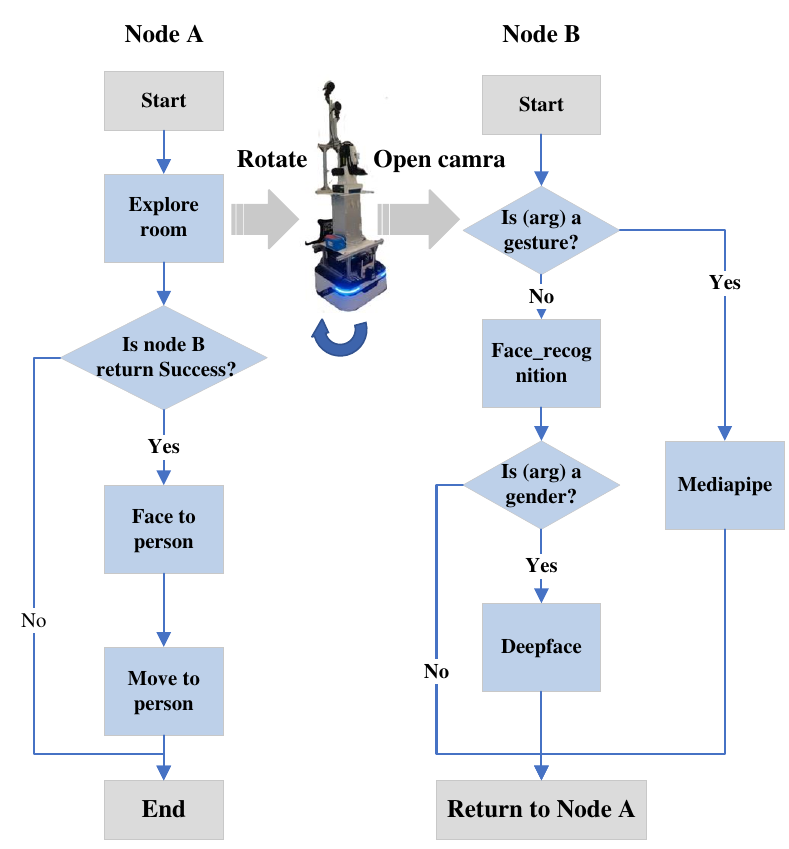}
  \caption{Flowchart of \textit{``look for person''}.}
  \label{fig2}
\end{figure}

\subsection{Generating State Machines with LLM}

We propose a task planning method $f_{prompt}(\cdot)$ based on a constrained LLM prompt scheme, which can generate an executable actions sequence from a command in natural language. An exceptional handling module is further proposed to deal with LLM hallucinationss problem~\cite{ji2023survey,huang2023survey}. With our proposed task planning method, the LLM can generate an executable action sequence in the form of a Pythonic list, which can be used to construct the SMACH~\cite{bohren2010smach} state machine.

\subsubsection{Constrained LLM Prompt}
Specifically, we define eight primitive actions: \textit{``move to''}, \textit{``look for obj''}, \textit{``look for person''}, \textit{``follow''}, \textit{``grasp''}, \textit{``pass to''}, \textit{``speak''}, and \textit{``answer''}, which can implement all commands generated by RoboCup@Home Command Generator~\cite{generator}.  Firstly, we provide these primitive actions to the LLM as Pythonic import statements, to ensure the generated action sequences are executable. For example, in the task \textit{Give me a cola}, the LLM may generate the first action ``[\textit{find, cola}]''. However, the robot might not define a primitive action to find or have already had a similar primitive action such as \textit{``look for obj''}. As shown in Fig. 1, we explicitly import executable primitive actions as Pythonic function structures. These encourage the LLM to constrain the task planning result to executable primitive actions that are predefined. And the available objects in the environment can be used to be the parameters of Pythonic action functions.

To facilitate the LLM has more precise comprehension of task nature and requirements, we provide a few parsing examples to LLM for few-shot learning.  These parsing examples are fully executable and show the LLM how to parse a given command using the predefined primitive actions and complete tasks in a subjectively reasonable manner. Due to token length limitations, we provide 23 parsing examples, covering a broad spectrum of commands generated by RoboCup@Home Command Generator\cite{generator}. For complex commands prone to misunderstanding, we provide multiple examples to enable the LLM to learn the task planning example accurately. For example, in the task \textit{Meet Jennifer at the sink, follow her, and take her back}, it is hard for the LLM to understand take her back and parse into the form ``[actions, parameters]''. Thus we interpret “back” as “initial location” and the command \textit{Meet Jennifer at the sink, follow her, and take her back} is represented as ``[[\textit{move to}, \textit{sink}], [\textit{look for person}, \textit{Jennifer}], [\textit{follow}, \textit{Jennifer}], [\textit{speak}, \textit{Please follow me!}], [\textit{move to}, \textit{initial location}]]''. 

Subsequently, we add parsing requirements, as shown in Fig. 1. In this way, we can guide the LLM in responding in the same format as the provided examples and help LLM better clarify commands that are easy to confuse. Then we add the current parsing task and input these sentences to the LLM. Due to the few-shot learning and explicit parsing requirements, the LLM finally generates an executable action sequence in the form of a Pythonic list, which can be used to construct the SMACH~\cite{bohren2010smach} state machine.

\subsubsection{Exceptional Handling}
We also employ an exceptional handling module to ensure the LLM-generated results are admissible in the current environment. Prior works introduce LLM hallucinationss~\cite{ji2023survey,huang2023survey}, resulting in content that is inconsistent with real-world facts or user inputs. For example, given a command \textit{Give me a cola} with the above prompt scheme, the task planning result could be ``[\textit{look for obj}(\textit{cola})]'' for faithfulness hallucinations which refers to the divergence of generated content from user instructions, and ``[\textit{find}, \textit{cola}]'' for factuality hallucinations which refers to factual inconsistency or fabrication because we never define the primitive action find. To deal with LLM hallucinations problem, we add new prompt instructions when we detect that the answer deviates from the specified format or the parsed function list contains actions beyond the provided primitive actions. We add a new prompt \textit{Please note the format of the answer!} and send a request to LLM again when the answer deviates from the specified format. If the parsed action list contains actions beyond the provided primitive actions, a prompt is added \textit{Please note that scheduled subtasks need to be used to complete task planning.} The robot will determine that the command cannot be parsed if an exception occurs five consecutive times.

\subsection{Implementation of Primitive Actions}
In this section, we delineate the defined primitive actions and furnish their specific details:

\textbf{1) \textit{``look for person''}}: Node A is the main node of \textit{``look for person''} function, which encompasses three sub-states: EXPLORE\_ROOM, FACE\_TO\_PERSON, and MOVE\_TO\_PERSON, designed for searching specific individuals in the environment. The functions of each state are as follows: 
\begin{itemize}
    \item EXPLORE\_ROOM: Involves room exploration, human detection, and identification of gesture characteristics.
    \item FACE\_TO\_PERSON: Directs the robot to face humans and confirm their gender.
    \item MOVE\_TO\_PERSON: Guides the robot to move in front of individuals meeting the specified criteria.
\end{itemize}
As shown in Fig 3, firstly, the robot rotates at a fixed point and begins to execute node B. Depending on the given parameter, the robot will then detect the gesture characteristics of people or facial characteristics. Finally, the detection results will be returned to Node A by the topic communication mechanism in ROS, for the robot executing subsequent states.

\textbf{2) \textit{``follow''}}: This action entails following the person in front of the robot, implemented through MediaPipe~\cite{lugaresi2019mediapipe} gesture recognition and skeleton tracking. Specifically, we utilize a RealSense depth camera for RGB images and depth maps. Then we detect the human gesture from the video frames, which determines subsequent states, including following, pausing, and termination. When the next state is following, we locate the coordinates of the midpoint of the human shoulders through skeleton tracking. By combining the coordinates and depth maps, we can obtain the PD control quantity for robot motion.

\textbf{3) Other Primitives}: \textit{``move to''} action aims to navigate the robot to the target position. We define the coordinates of different locations in the environment and use the move\_base package in ROS for path planning.

\textit{answer} action aims to respond to questions from individuals. We employ the LLM to generate a response to a question. The robot initiates the \textit{``answer''} action, prompting a new round of Q\&A. Parsing requirements are adjusted to \textit{Please answer the following questions in English in no more than 30 words.} After obtaining the answer's text information, the \textit{``speak''} action is employed to respond to the questioner. The Q\&A module incorporates a mechanism to determine the inability to answer after 5 consecutive failed attempts.

\section{Experiments}
In this section, we first evaluate the robot’s plan decomposition ability for tasks expressed in spoken language. Then
we apply our method to a robot in the real world and selected
four representative commands containing different primitives for testing. Finally, we design an evaluation metric to
compare the performance of different large models on plan
decomposition.
\subsection{Evaluation of Robot's Plan Decomposition}
\begin{figure}[t]
  \centering
  \includegraphics[width=9cm]{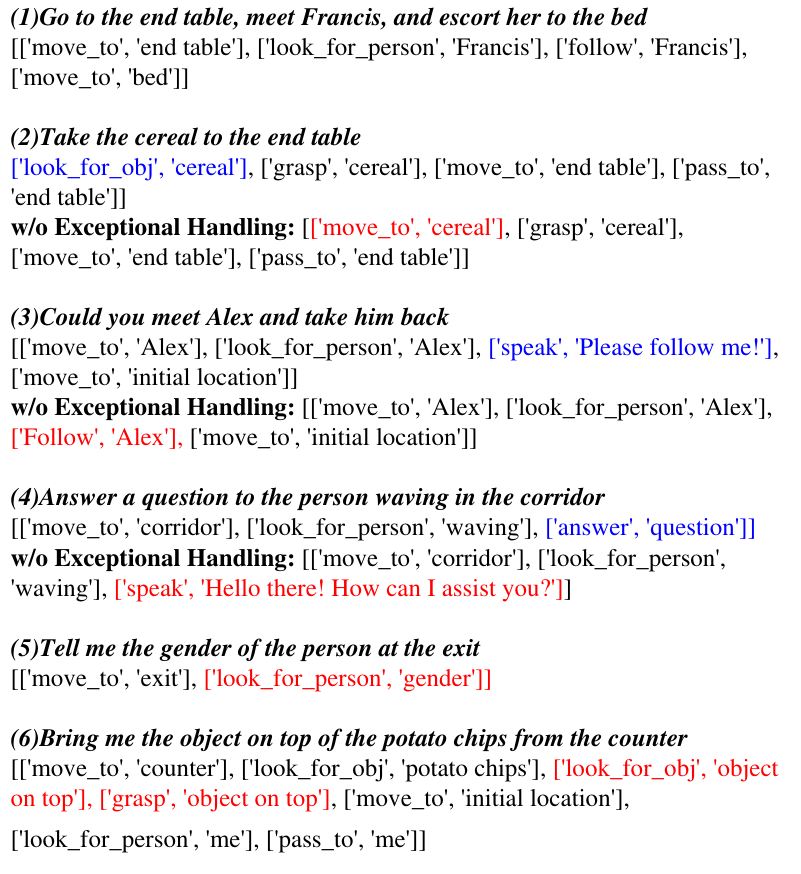}
  \caption{The typical commands and test results.}
  \label{fig2}
\end{figure}

\begin{figure*}[htbp]
  \centering
  \includegraphics[width=16cm]{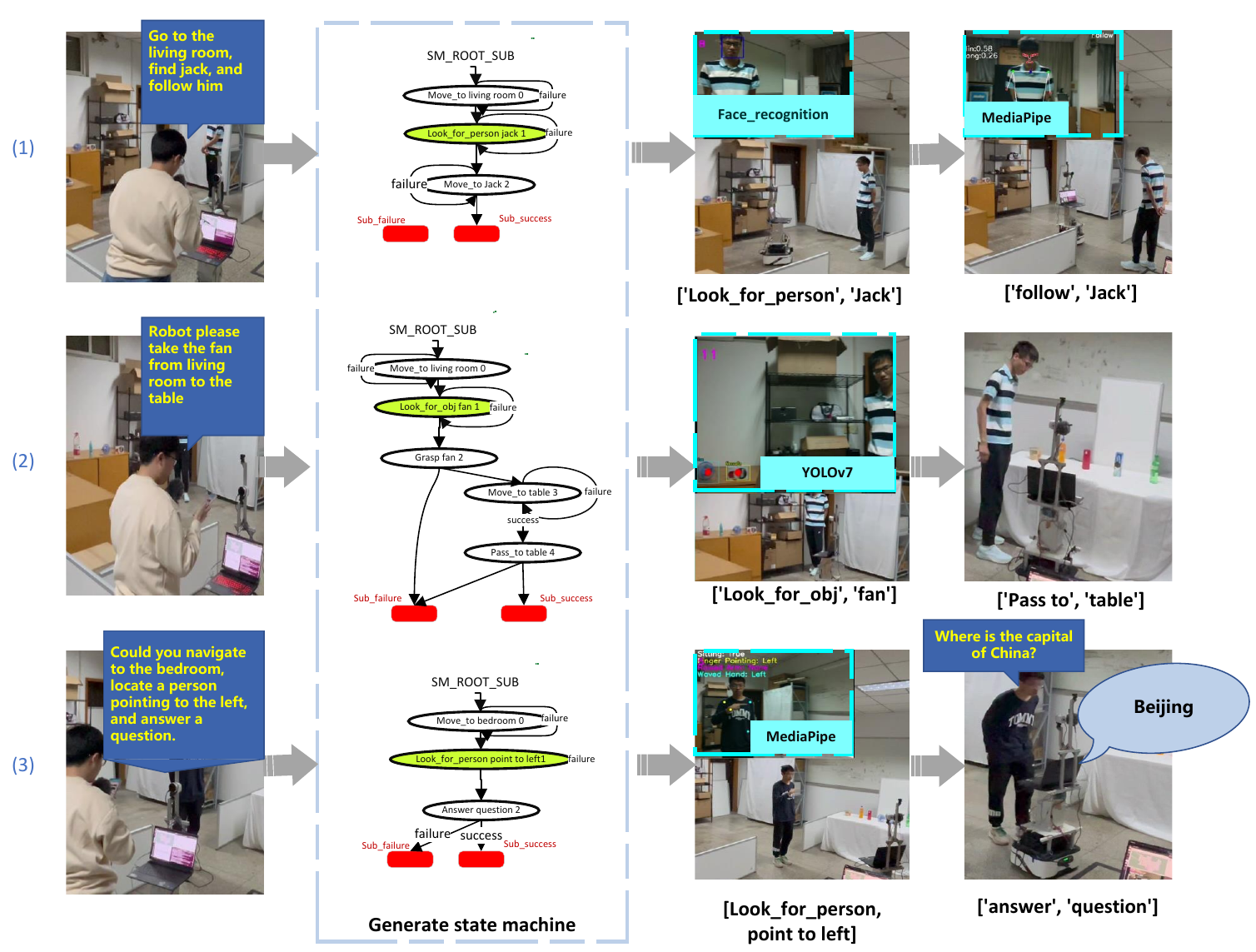}
  \caption{The process of executing commands in spoken language.}
  \label{fig3}
\end{figure*}
 In order to study the effectiveness of our designed prompts in Sec 3.2, we test 100 commands generated by RoboCup@Home Command Generator~\cite{generator}. To evaluate the correctness and executability of generated action plans, we conduct human evaluations. Among the results, 83 tasks can output be decomposed to the correct order of primitive functions with corresponding parameters and 69 of the tasks can be completely executed with the current implementation of primitive functions. This result indicates that the prompted LLM can generate correct and realistic action plans in most cases. 
 
Fig 4 shows 6 typical examples, in which the black text shows the commands and generated sequences, the red text shows the incorrect parts and the blurred text shows the improvement made by our method. Commands similar to (1) (mainly about navigation and following) account for the highest proportion of the output of the instruction generator, and our method has demonstrated excellent performance on such commands. As shown in (2), our method can avoid generating function \textit{``grasp''} without \textit{``look for obj''}. Notably, for voice interaction commands like (3) and (4), method in\cite{obinata2023foundation} may generate wrong plans that confuse the robot itself and objects (e.g. ``[\textit{speak}, \textit{Please follow me!}]'' in (3) may be mistaken for ``[\textit{follow}, \textit{Alex}]'' and ``[\textit{answer}, \textit{question}]'' in (4) may be mistaken for ``[\textit{speak}, \textit{Can I help you}]''), while our prompts scheme with exceptional handling can avoid this. In cases (5) and (6), there are no incorrect primitive functions but the sequences can not be executed completely because the semantic information for the objects was regarded as a part of the objects’ names, resulting in the failure during the execution process.

\subsection{Real-world Robotic Experiment}
We evaluate our proposed method on a real-world robot platform designed to resemble a family scene containing furniture like a bed, bookstore, tea table, and so on. The setup consists of an AgileX Robotics TRACER MINI equipped with a HOKUYO URG-04LX-UG01 lidar and an Intel RealSense RGB-D camera looking straight ahead at the robot. As described in Sec 3.3, aside from pretraining of YOLOv7~\cite{wang2023yolov7}, the system does not require any model finetuning to perform general purpose service. 

Here we take three representative commands to test the performance of the proposed method on the real-world robot. The commands consist of different combination of primitives: 1) \textit{``navigate''} + \textit{``look for obj''} +\textit{``grasp''}, 2) \textit{``navigate''} + \textit{``look for person''}(gender) + \textit{``speak''}, 3) \textit{``navigate''} +  \textit{``look for person''}(gesture)+ \textit{``answer question''}. Fig 5 shows the execution process of the selected commands. Take task 3) For example, a person tells the robot “Could you navigate to the bedroom, locate a person pointing to the left, and answer a question ?”. Then the robot generates a SMACH based on the output of the LLM——"[[\textit{move to}, \textit{bedroom}], [\textit{look for person}, \textit{point to the left}], [\textit{answer}, \textit{question}]]". Based on the generated SMACH~\cite{bohren2010smach}, the robot starts to move. First, the “bedroom” has been registered as a navigation point, so the robot navigates to it. Then the robot keeps spinning around and invoking MediaPipe at the same time, until finding a person pointing to the left. Finally, the robot will stop in front of the person and wait for the question. The processes of other tasks are similar to task 3), during which the robot can finish all the given tasks, indicating a promising performance of our method on general purpose service robot.

\subsection{Comparison of State Generation Performance for LLMs}
\begin{table}[!htb]
  \centering
  \caption{performance comparison of LLMs}
  \scalebox{0.8}{
  \begin{tabular}{c|ccccc}
    \hhline
    commands & Spark\cite{Spark}&llama\cite{llama}&TigerBot\cite{tigerbot}&ERNIE-Bot\cite{wenxin}\\ \hline
    Type A& 96\%&88\%&76\%&98\%\\ 
     Type B&63\% &85\%&90\%&93\%\\ 
    Type C &51\% & 69\%&70\%&78\%\\ 
    \hhline
  \end{tabular}
  }
\end{table}

In this section, we study the effect of using different LLMs for the GPSR task. We compare four large language models, including Spark v2.0\cite{Spark}, ERNIE-Bot 4.0\cite{wenxin}, Tigerbot\cite{tigerbot}, and llama2\_70b\cite{llama}. We generate 100 commands by RoboCup@Home Command Generator\cite{generator} for comparison. To study the LLMs' performance for tasks containing different primitives, we divide the evaluation set into three types: (A) navigate to somewhere, look for a person, and follow, (B)tasks requiring the robot to look for an object and pass, (C)tasks requiring the robot to speak or answer someone's question. We assess the accuracy of each model for each task category. The accuracy refers to the percentage of correctly parsed primitive functions with parameters compared to the correct state machine sequence. The results are shown in Table 1.

For type A, all LLMs exhibit good performance, with ERNIE-Bot 4.0\cite{wenxin} achieving nearly perfect results. For type B, Spark v2.0\cite{Spark} performs poorly, while the other models show no significant differences. Notably, none of the models perform well in type C. We hypothesize that this is because the questions to be answered are mostly related to visual information, which large language models can’t handle effectively. Among the four tested models, ERNIE-Bot 4.0\cite{wenxin} shows outstanding performance in decomposing both easy and difficult tasks compared to other models, hence we ultimately select the ERNIE-Bot 4.0\cite{wenxin}.

\section{Conclusion and Future Work}
 In this paper, we propose a task planning method based on a constrained LLM prompt scheme, which can generate an executable action sequence from a command. An exceptional handling module is further proposed to deal with the LLM hallucinations problem. LLM hallucinations could result in content that is inconsistent with real-world facts or user inputs, including faithfulness hallucinations and factuality hallucinations. We add new prompt instructions when we detect that the answer deviates from the specified format or the parsed function list contains actions beyond the provided primitive actions. This module can ensure the LLM-generated results are admissible in the current environment. We evaluate our method on commands generated by the RoboCup@Home Command Generator\cite{generator}, observing that the robot demonstrates exceptional performance in both comprehending instructions and executing tasks.

 While we present a viable way to employ LLM for robot task planning and deal with LLM hallucinations issues, it has limits. First, for the instruction sets that are more semantically rich and abstract, such as ALFRED{\cite{shridhar2020alfred}} and BEHAVIOR{\cite{srivastava2022behavior}}, our defined primitive actions may not be executable due to the lack of specific individuals or objects as arguments. In future work, Visual Language Models could be introduced to extract additional semantic information from the environment, thus enhancing the robot's generalization capabilities. Secondly, our exceptional handling module can take effect after the occurrence of LLM hallucinations issues, yet its capability remains limited, requiring deeper research on designing more effective prompts to reduce the occurrence of illusion problems.
\balance

\bibliographystyle{ieeetr}  
\bibliography{reference} 

\end{document}